\documentclass[10pt,twocolumn,letterpaper]{article}

\usepackage[pagenumbers]{cvpr} 
\usepackage{enumitem}
\usepackage{graphicx}
\usepackage{amsmath}
\usepackage{amssymb}
\usepackage{booktabs}
\usepackage{multirow}

\usepackage{algorithm}
\usepackage{algpseudocode}

\definecolor{cvprblue}{rgb}{0.21,0.49,0.74}
\usepackage[pagebackref,breaklinks,colorlinks,allcolors=cvprblue]{hyperref}

\def\confName{CVPR}
\def\confYear{2025}

\title{Tuning-Free Long Video Generation via Global-Local Collaborative Diffusion}

\author{%
  Yongjia Ma$^1$\qquad 
  Junlin Chen$^2$ \qquad  
  Donglin Di$^1$ \qquad     
  Qi Xie$^3$\qquad\\[1ex]
    Lei Fan$^4$\qquad
    Wei Chen$^1$\qquad
    Xiaofei Gou$^{1}$\qquad
    Na Zhao$^{5}$\qquad
    Xun Yang$^3$\\[1ex]
  \\
    $^1$Space AI, Li Auto\quad
    $^2$Zhejiang University\quad
    $^3$University of Science and Technology of China\\
    $^4$University of New South Wales\quad
    $^5$Singapore University of Technology and Design
}

\begin{document}

\vspace{-0.8cm}

\twocolumn[{
\renewcommand\twocolumn[1][]{#1}
\maketitle

\begin{center}
    \centering
    \captionsetup{type=figure}
    \includegraphics[width=\textwidth]{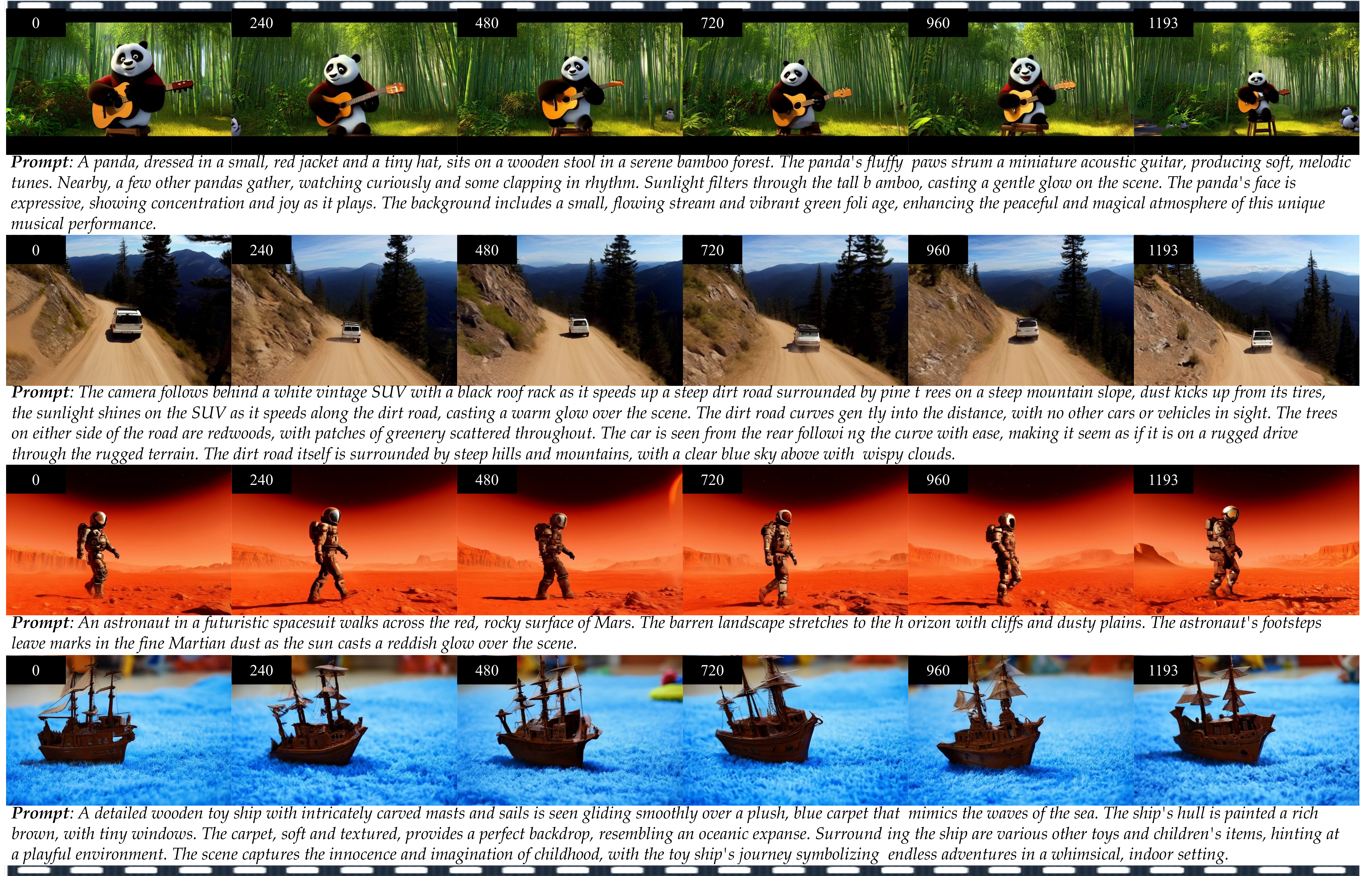}
       \caption{
Demonstration of a long video exceeding 1,000 frames generated by our GLC-Diffusion based on CogVideoX \cite{yang2024cogvideox}.  Each frame index is displayed in the top-left corner of the images. Our method enables the production of high-quality, extended-length videos from models initially trained on short clips (\textit{e.g.}, 49 frames). 
We present video results that are 25 $\times$ longer than the original clips, highlighting the scalability of our approach.
GLC-Diffusion effectively maintains global content consistency and enhances local temporal coherence throughout the video without additional training.
       }  
    \label{fig:fig1_showcase}
\end{center}%
}]

\maketitle
\vspace{-0.8cm}

\begin{abstract}

\vspace{-12pt}

Creating high-fidelity, coherent long videos is a sought-after aspiration. While recent video diffusion models have shown promising potential, they still grapple  with spatiotemporal inconsistencies and high computational resource demands. We propose GLC-Diffusion, a tuning-free method for long video generation. It models the long video denoising process by establishing denoising trajectories through Global-Local Collaborative Denoising to ensure overall content consistency and temporal coherence between frames. Additionally, we introduce a Noise Reinitialization strategy which combines local noise shuffling with frequency fusion to improve global content consistency and visual diversity. Further, we propose a Video Motion Consistency Refinement (VMCR) module that computes the gradient of pixel-wise and frequency-wise losses to enhance visual consistency and temporal smoothness. 
Extensive experiments, including quantitative and qualitative evaluations on videos of varying lengths (\textit{e.g.}, 3× and 6× longer), demonstrate that our method effectively integrates with existing video diffusion models, producing coherent, high-fidelity long videos superior to previous approaches.

\end{abstract}    
\vspace{-15pt}

\section{Introduction}
\label{sec:intro}
\vspace{-3pt}




Diffusion Models (DMs) have revolutionized the field of image synthesis \cite{ho2020denoising,rombach2022high,podell2024sdxl,esser2024scaling,10447068,pmlr-v202-bar-tal23a}. Building upon this success, there has been a growing interest in extending these capabilities to video generation \cite{li2023videogen,wang2023lavie,ma2024latte,ho2022video,xing2023survey,yang2024cogvideox}. Video generation has a significant impact on various applications, including film production, game development, and artistic creation \cite{liu2024sora,xing2023survey,yang2024tv}. 
Compared to image generation \cite{wang2023internvid,long2024videodrafter,esser2023structure}, video generation demands significantly greater data scale and computational resources due to the high-dimensional nature of video data.
This necessitates a trade-off between limited resources and model performance for Video Diffusion Models (VDMs) \cite{li2024training,menapace2024snap,wu2025freeinit}.

The inherent multidimensional complexity of long videos poses significant challenges under existing resource constraints \cite{li2024survey,wu2023lamp,menapace2024snap,chen2023videodreamer}.
Recent VDMs are typically trained on a limited number of video frames, which restricts their generative capacity to producing videos of only a few seconds in length \cite{li2024training,menapace2024snap}. 
Some studies \cite{henschel2024streamingt2v,duan2024exvideo,wang2024loong} enhance long video generation capabilities by designing additional learnable models. However, these approaches require substantial computational resources and large datasets and are difficult to be compatible with existing different VDMs \cite{yu2024efficient,lv2024fastercache,yoon2024safree}.
Additionally, other methods extend long video generation capabilities in a tuning-free manner by stitching together video clips with sliding windows in the latent space or temporal attention mechanisms \cite{wang2023gen,qiu2024freenoise,oh2025mevg,lu2024freelong}.
However, these methods only consider local denoising, resulting in persistent issues like frame skipping, motion discontinuity, and content inconsistency, which fail to produce to meet the pursuit of high-fidelity long video generation \cite{li2024survey}.

In this paper, we propose GLC-Diffusion, a tuning-free method for creating high-quality, coherent long videos. The core idea is to model the denoising process as a unified optimization problem \cite{pmlr-v202-bar-tal23a,wang2023gen} using a dual-path Global-Local Collaborative Denoising (GLCD) mechanism, where the denoising trajectory is partitioned into global and local paths. Specifically, In the global path, Global Dilated Sampling is employed to capture long-range temporal dependencies, preserving overarching scene consistency and continuity throughout the video. In the local path, a Local Random Shifting Sampling is introduced to apply randomly shifted overlapping denoising windows across timesteps, which corrects seams and temporal artifacts at a given timestep during subsequent steps, strengthening local temporal coherence and smoothing frame transitions.

To further enhance video generation quality, we introduce a Noise Reinitialization strategy that combines local noise shuffling with frequency fusion to address the motion diversity limitations inherent in FreeNoise \cite{qiu2024freenoise}. It enhances both temporal consistency and visual diversity in the generated videos.
Following the dual-path output, we additionally introduce the Video Motion Consistency Refinement module, which refines latent variables through gradient descent at each denoising step. It aligns predicted latent motion vectors by minimizing a composite loss function that incorporates both pixel-wise and frequency-wise losses, thereby optimizing visual consistency and temporal smoothness across frames.

To validate our approach, we conducted extensive experiments based on the CogVideoX \cite{yang2024cogvideox} model, extending its generative capacity from 48 frames to over 1,000 frames, as illustrated in Figure.~\ref{fig:fig1_showcase}. Our method serves as a plug-and-play component within existing video diffusion frameworks, significantly enhancing the temporal coherence and visual fidelity of generated long videos, as demonstrated in both quantitative and qualitative evaluations. Our contributions are summarized as follows:

\begin{itemize}
\item We propose the Global-Local Collaborative Denoising (GLCD) mechanism, modeling the long video denoising process as a unified optimization problem that integrates global and local denoising paths to enhance both content consistency and temporal coherence without requiring additional training.
\item We introduce the Noise Reinitializatio strategy, which balances long-term temporal correlation and diversity, effectively improving global content consistency in the generated videos.
\item We develop the Video Motion Consistency Refinement (VMCR) module, refining latent variables through gradient descent to further enhance visual consistency and temporal smoothness across frames.
\item Our approach significantly extends the frame generation capacity of pretrained models like CogVideoX, outperforming previous methods in terms of coherence and fidelity.
\end{itemize}

\begin{figure*}[tpbh]
  \centering
\includegraphics[width=\textwidth]{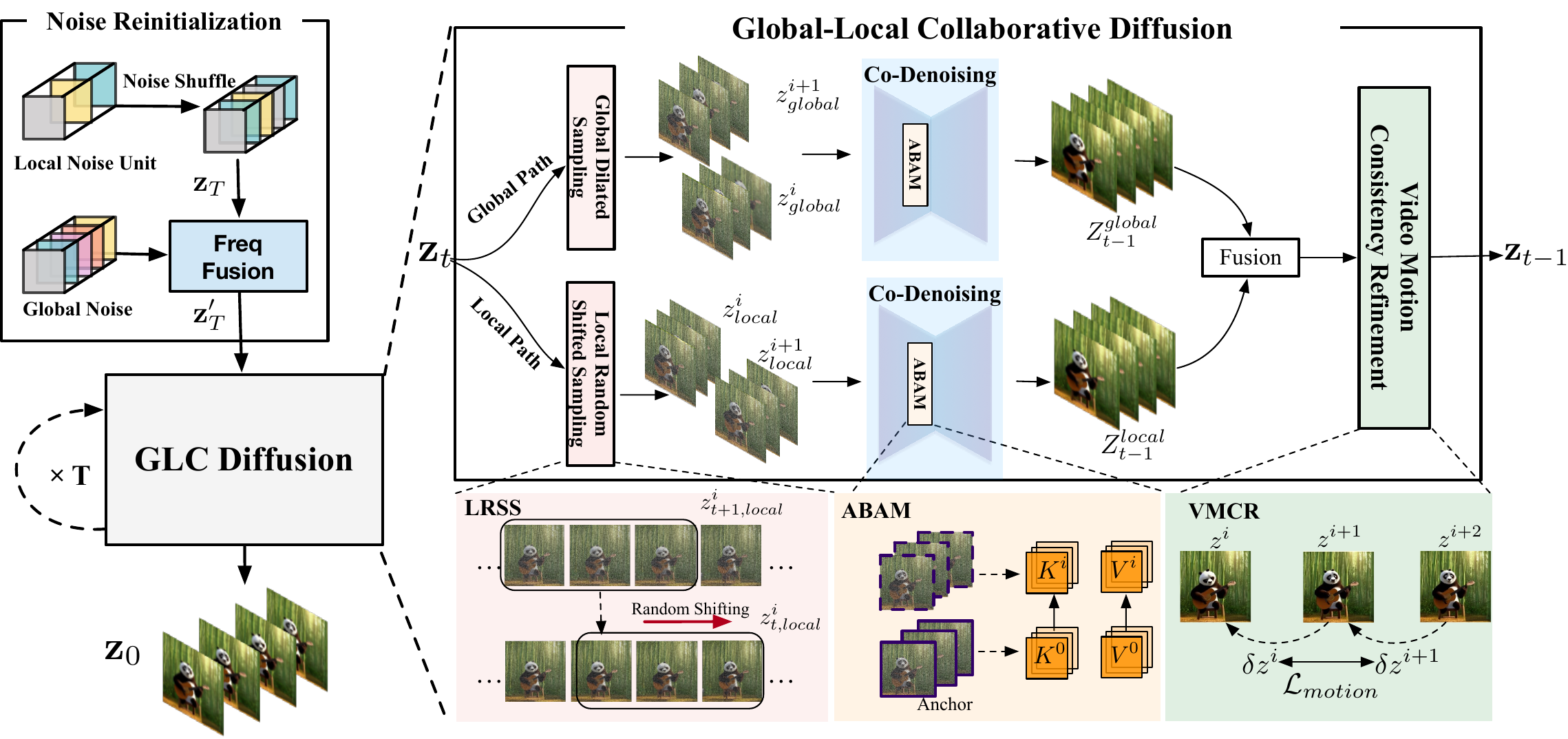}
   \caption{Overview of our GLC Diffusion. It illustrates the denoising process from \( z_t \) to \( z_{t-1} \), integrating our proposed modules: Global-Local Collaborative Denoising (GLCD), Local Random Shifting Sampling (LRSS), Attention-Based Adaptive Modulation (ABAM), and Video Motion Consistency Regularization (VMCR). GLCD consists of global and local denoising paths to maintain overall content consistency and enhance local temporal coherence. LRSS improves spatio-temporal coherence by sampling local frames with random shifts. ABAM adaptively modulates attention weights to emphasize important regions, while VMCR enforces motion consistency across frames.
}
   \label{fig:method}
\end{figure*}

\vspace{-6pt}

\section{Related Work}
\label{sec:relatedwork}
\subsection{Video Diffusion Models}

Video diffusion models build on the success of diffusion models in image generation, extending them into the temporal dimension for video generation \cite{he2022latent}. 
By integrating a 1D temporal convolution layer into the traditional 2D U-Net, these methods \cite{singer2023makeavideo,blattmann2023stable,guo2024animatediff} aim to emulate 3D convolution effects, leveraging image-text pair training and temporal context learning to connect videos with textual descriptions. Additionally, Sora \cite{brooks2024video} initiates the new era of video generation utilizing DiT-based architecture \cite{Peebles_2023_ICCV}. Video DiTs \cite{ma2024latte,yang2024cogvideox,pku_yuan_lab_and_tuzhan_ai_etc_2024_10948109,opensora} significantly enhances the model's ability to capture long-term dependencies, thereby improving the quality and diversity of generated videos. However, current models face two main challenges: computational resource limitations restrict processing to short video segments, and insufficient training data to support long video generation. In this paper, we propose a plug-and-play, tuning-free method that extends existing video diffusion models to generate longer and more consistent videos.

\vspace{-3pt}

\subsection{Long Video Generation}

Recent research has explored tuning-free long video generation using short video diffusion models \cite{zhang2024tvg,qiu2024freetraj,li2024training,henschel2024streamingt2v}. Gen-L-Video \cite{wang2023gen} extends videos by combining overlapping sub-sequences with a sliding window method during the denoising process. FreeNoise \cite{qiu2024freenoise} adopts sliding window temporal attention and noise initialization strategies to maintain temporal consistency. FIFO-Diffusion \cite{kim2024fifo} proposes latent partitioning and lookahead denoising to generate infinitely long videos. MEVG \cite{oh2025mevg} 
introduces techniques such as dynamic noise, last-frame-aware inversion, and structure-guided sampling to generate long videos with temporal continuity under multi-text conditions. FreeLong \cite{lu2024freelong} employs the SpectralBlend Temporal Attention mechanism to fuse global low-frequency features with local high-frequency features, enhancing the consistency and fidelity of long video generation. However, these methods have issues with long-term inconsistency in video generation, making it difficult to maintain spatiotemporal continuity. 
To mitigate these, we construct a unified optimization framework that incorporates a global-local collaborative denoising path, effectively enhancing both content consistency and temporal coherence in long video generation.

\vspace{-3pt}

\section{Methodology}
\label{sec:Methodology}

\subsection{Preliminary}

Initially, we introduce a pre-trained diffusion model, denoted as $\Phi$, which operates within the latent space $\mathbf{z} = \mathbb{R}^{t\times h \times w \times c}$  under the condition of  input  $y$ . Deterministic DDIM sampling \cite{song2021denoising} is employed for inference:
\vspace{-3pt}
\begin{equation}
    z_{t-1}=\sqrt{\frac{\alpha_{t-1}}{\alpha_t}}z_t+\left(\sqrt{\frac{1}{\alpha_{t-1}}-1}-\sqrt{\frac{1}{\alpha_t}-1}\right)\cdot\Phi(z_t,t,y),
    \label{eq:ddim}
\end{equation}
where $z_t \in Z$ and $\alpha_t$ are determined by the DDIM schedule ${\beta_i \mid i = 1, 2, \ldots, T, \beta_i \in (0, 1)}$. After $T$ denoising steps, we obtain the image $z_0$ from the initial Gaussian noise $z_T$:

In Gen-L-Video \cite{wang2023gen}, a set of mapping relations $F_i$ is defined to project all original videos $z_t$ in the denoising trajectory onto short video clips $z_i^t$:
\begin{equation}
    z_t^i=F_i(z_t),
\end{equation}

Each video clip is independently denoised using $f_\Phi$ as proposed in Eq.~\ref{eq:ddim}: $z^{i}_{t-1} = f_\Phi(z^{i}_{t}, t,  y)$. Then, based on the manifold hypothesis, a global least squares optimization is established to minimize the discrepancy between each clip $F_i(z_{t-1})$ and its denoised counterpart $f_\Phi(F_i(z_{t}), t,  y)$. This process merges different crops into a single long video $z'$. Due to the properties of $\Phi$, this optimization problem has a unique solution:
\vspace{-3pt}
\begin{equation}
    z_{t-1}'=\frac{\sum_iW_i\otimes F_i^{-1}(f_\Phi(z_t^i,t, y))}{\sum_iW_i},
    \label{eq:solver}
\end{equation}
where $W_i$ represents the pixel weights of the video clip $v_i^t$, and $\otimes$ denotes the tensor product.

\subsection{Global-Local Collaborative Denoising}

Global-Local Collaborative Denoising (GLCD) is proposed to establish denoising trajectories for long videos, incorporating both global and local denoising paths, as illustrated in Figure.~\ref{fig:method}.

\vspace{-6pt}

\paragraph{Global Path}
We aim to maintain the overall content consistency of the video by capturing long-range temporal dependencies. This is achieved through Global Dilated Sampling, which involves padding the latent variable $\boldsymbol{z}_t$ and sampling frames at equal intervals to create global video clips. The padding ensures that boundary frames are adequately represented, preventing edge artifacts. Specifically, we define the mapping function $F_{\mathrm{global}}^i$ as:
\begin{equation}
F_{\mathrm{global}}^i(\boldsymbol{z}_{t}) = \boldsymbol{z}_t[ s_i + d\times j ], 
\end{equation}
where $s_i $ represents the starting frame index of the $ i $-th video clip, $ d $ is the dilation rate determining the sampling interval, $ j = 0, 1, \ldots, L - 1 $ indicates the $j$-th video clip, $ L $ denotes the video clip length, and $ \boldsymbol{z}_t[n] $ represents the $ n $-th frame of the padded latents.

By applying dilated sampling, we construct a series of global latent representations ${Z}_{t}^{\mathrm{global}} $ :
\begin{equation}{Z}_{t}^{\mathrm{global}} = F_{\mathrm{global}}(\boldsymbol{z}_{t})=[\boldsymbol{z}_{t,\mathrm{global}}^0, \boldsymbol{z}_{t,\mathrm{global}}^1, \ldots, \boldsymbol{z}_{t,\mathrm{global}}^{N-1}],
\end{equation}

It allows the model to focus on the overarching themes and narratives during the denoising process, effectively preserving global content consistency across the entire video. By capturing the long-range temporal structures, the global denoising path ensures that the generated video maintains coherence in terms of story, characters, and settings.

\vspace{-10pt}

\paragraph{Local Path}
Our goal is to enhance local temporal coherence by correcting inter-frame discontinuities. This is achieved through a Local Random Shifting Sampling strategy. At each timestep $ t $, we partition the video into overlapping local clips of fixed length $ L $ and apply a random temporal shift $ s_i^t $ to the starting point of each clip. We define the mapping function $ F_{\mathrm{local}}^{i,t} $ as:
\vspace{-6pt}
\begin{equation}
F_{\mathrm{local}}^{i,t}(\boldsymbol{z}_t) = \boldsymbol{z}_t[ s_i^t + j ],
\end{equation}
where $ j = 0, 1, \ldots, L - 1 $, 
$s_i^t $ represents the starting frame index of the $ i $-th video clip with random temporal shift  at timestep $t$,
and $ \boldsymbol{z}_t[n] $ represents the $n $-th frame of the latent variable, as defined in the Global Path.

The overlapping nature of the clips, combined with the random shifts, encourages the model to consider different temporal contexts during denoising. This results in a series of local latent representations:
\vspace{-6pt}
\begin{equation}
{Z}_{t}^{\mathrm{local}} = F_{\mathrm{local}}^t(\boldsymbol{z}_{t})=[\boldsymbol{z}_{t,\mathrm{local}}^0, \boldsymbol{z}_{t,\mathrm{local}}^1, \ldots, \boldsymbol{z}_{t,\mathrm{local}}^{M-1}],
\end{equation}

By incorporating random temporal shifts, we enhance the diversity of temporal relationships captured by the model, which effectively corrects inter-frame discontinuities and reduces temporal artifacts such as flickering or abrupt changes. This leads to smoother frame transitions and a more coherent and seamless video output.

\vspace{-10pt}

\paragraph{Unified Optimization Framework}

To effectively integrate the global and local denoising paths, we embed our GLCD method into a unified optimization framework.

We define the total number of video clips as \( K = N + M \), where \( N \) and \( M \) represent the numbers of global and local video clips, respectively. Using a unified index \( k = 0, 1, \ldots, K - 1 \), the mapping function \( F_k(\boldsymbol{z}_t) \), the weight matrix \( W_k \), and the reconstructed variable \( \boldsymbol{z}_k \) are defined based on whether \( k \) falls within the range of global or local clips. For \( k < N \), corresponding to the global clips, \( F_k(\boldsymbol{z}_t) \) applies the global mapping function, \( W_k \) scales with the annealing coefficient \( \sqrt{\gamma} \). When \( k \geq N \), for local clips, the local mapping function \( F_{k - N}^{\text{local}}(\boldsymbol{z}_t) \) is used, with \( W_k \) scaled by \( \sqrt{1 - \gamma} \). This unified index allows consistent treatment across global and local paths, balancing their contributions through the coefficient \( \gamma \) in \( W_k \), and reconstructing each path’s latent variable according to its respective mappings.

Using these definitions, the long video denoising process is formulated as a single optimization problem \cite{pmlr-v202-bar-tal23a}:
\begin{equation}
\boldsymbol{z}_{t-1}=\arg\min_{z}\sum_{k=0}^{K-1}\left\|W_{k}\otimes\left(F_{k}(\boldsymbol{z})-\boldsymbol{z}^k_{t-1}\right)\right\|_{2}^{2}, 
\end{equation}
where $\boldsymbol{z}^k_{t-1}$ denotes $k$-th video clip for each path.

This optimization problem seeks the latent variable \( \boldsymbol{z}_{t-1} \) at timestep \( t - 1 \) that minimizes the weighted sum of reconstruction errors over all video clips from both global and local paths. It is essentially a weighted least squares formulation, a convex optimization problem that guarantees a unique global minimum.

Based on the manifold hypothesis \cite{pmlr-v202-bar-tal23a,lee2023syncdiffusion,zhou2024twindiffusion}, solving this optimization problem yields the optimal latent variable at timestep $t-1$:
\vspace{-3pt}
\begin{equation}
    \boldsymbol{z}_{t-1}=\gamma\times\mathcal{T}_{\mathrm{global}}(Z_{t-1}^{\mathrm{global}})+(1-\gamma)\times\mathcal{T}_{\mathrm{local}}(Z_{t-1}^{\mathrm{local}}),
\end{equation}
where $\gamma=\gamma_0*e^{\beta\times t}$, the annealing coefficient $c$ varies with the timestep. 
$\mathcal{T}$ denotes the calculation from Eq.~\ref{eq:solver} applied to each path. 
This closed-form solution demonstrates how the global and local denoising paths are harmoniously integrated. The annealing coefficient $\gamma$ dynamically balances the influence of each path during the denoising process, ensuring that both global consistency and local coherence are maintained throughout. 

\vspace{-12pt}

\paragraph{Anchor-Based Attention Mechanism}
To enhance temporal coherence within the diffusion model, we propose an Anchor-Based Attention Mechanism (ABAM) to replace the native attention mechanisms. Our method is designed to be compatible with models utilizing self-attention, ensuring broad applicability.

In denoising path, we consider the first video clip $\mathbf{v}_0 $ as the anchor clip. For any other clip $ \mathbf{v}_i $ ($ i = 1, 2, \ldots, N-1 $), we inject the key and value representations from the anchor clip into the attention computations of the current clip.
The key $ \mathbf{K}_{i,j} $ and value $ \mathbf{V}_{i,j}$ for frame $ j $ in clip $ i $ are updated as:
\begin{equation}
\begin{cases}
\mathbf{K}_{i,j} &=  \lambda \cdot \mathbf{K}_{i,j}^{\text{original}} + (1 - \lambda) \cdot \mathbf{K}_{\text{anchor}}, \\
\mathbf{V}_{i,j} &= \lambda \cdot  \mathbf{V}_{i,j}^{\text{original}} + (1 - \lambda) \cdot  \mathbf{V}_{\text{anchor}},
\end{cases}    
\end{equation}
where $ \mathbf{K}_{\text{anchor}} $ and $ \mathbf{V}_{\text{anchor}} $ are the Key and value representations of the anchor, $\lambda$ is a scaling factor controlling the influence of the anchor. The output of the attention mechanism for frame \( j \) in clip \( i \) is then: $    \mathbf{Att}_{i,j} = \text{softmax}( \mathbf{Q}_{i,j} \mathbf{K}_{i,j}^\top / {\sqrt{d_k}})\mathbf{V}_{i,j} $, where \( d_k \) is the dimensionality of the key vectors.

\subsection{Noise Reinitialization} 
To further enhance video generation quality, we introduce a Noise Reinitialization strategy that combines Local Noise Shuffle with frequency fusion. Inspired by \cite{qiu2024freenoise,wu2025freeinit}, we first utilize Local Noise Shuffling to generate the initial latent variables, $\boldsymbol{z}_T = \operatorname{shuffle}(\epsilon)$,  by denotes the operation of randomly shuffling the noise sequence within a local region. To enhance the motion diversity, we decompose the noise latent $\boldsymbol{z}_T$ into low and high-frequency components using a spatio-temporal frequency filter. Specifically, we combine the low-frequency content of \( \boldsymbol{z}_T \) with the high-frequency components of a  global Gaussian noise \( \eta \). The reinitialized noise latent \( \boldsymbol{z}'_T \) is computed as follows:

\begin{equation}
\begin{cases}
F_L(\boldsymbol{z}_T) &= \text{FFT3D}(\boldsymbol{z}_T) \odot H, \\
F_H(\eta) &= \text{FFT3D}(\eta) \odot (1 - H), \\
\boldsymbol{z}'_T &= \text{IFFT3D}(F_L(\boldsymbol{z}_T) + F_H(\eta)),
\end{cases}
\end{equation}
where \( \text{FFT3D} \) and \( \text{IFFT3D} \) are the 3D Fast Fourier Transform and its inverse, applied over both spatial and temporal dimensions. \( H \) represents a spatial-temporal low-pass filter (LPF), ensuring that low-frequency information is retained from $\boldsymbol{z}_T$, while the high-frequency randomness from \( \eta \) enhances visual details. The resulting $\boldsymbol{z}'_T$ serves as the initialized noise for subsequent DDIM sampling, contributing to improved frame quality and temporal alignment.

\begin{figure}[t]
  \centering
    \includegraphics[width=0.47\textwidth]{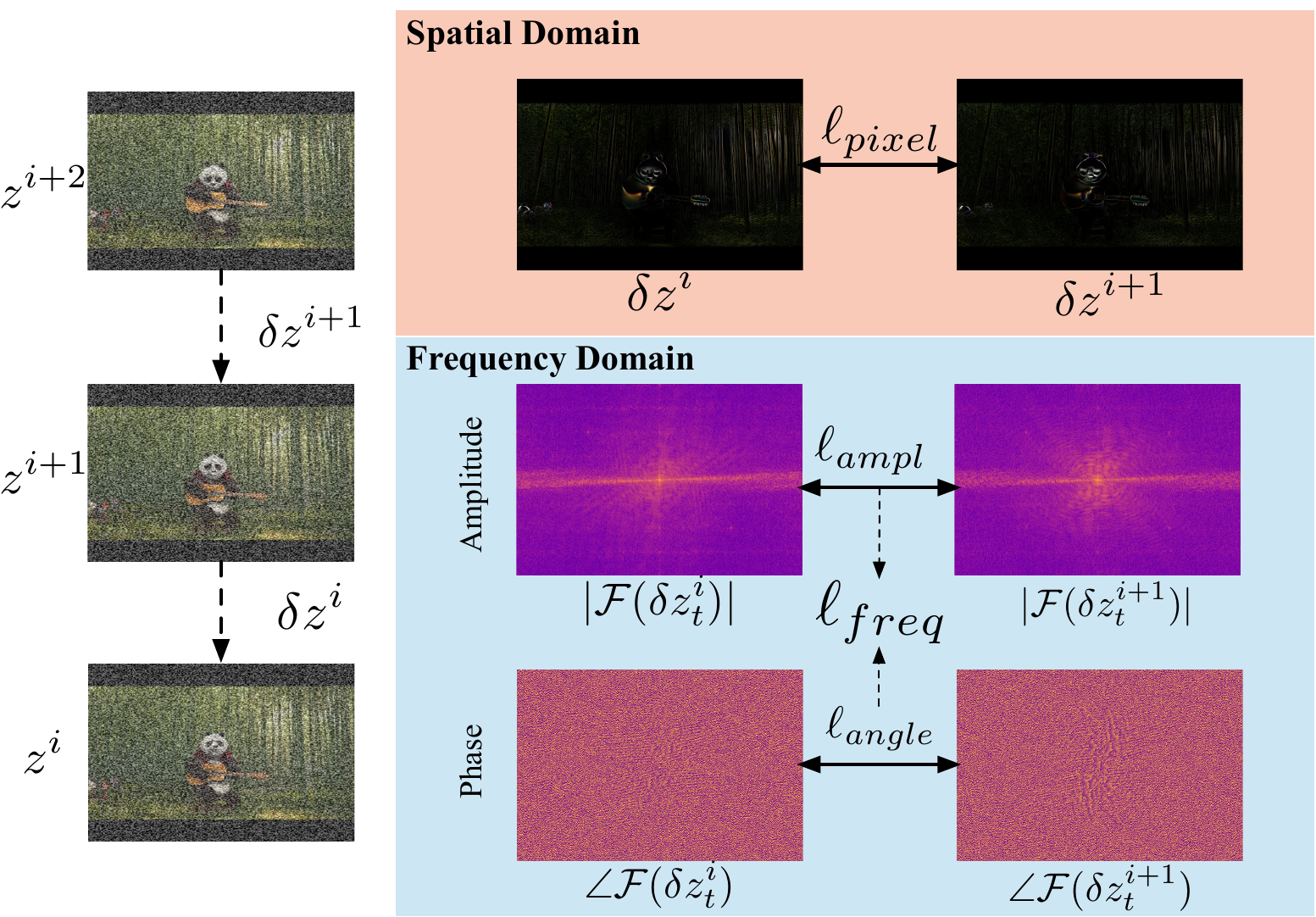}
    \vspace{-6pt}
   \caption{Illustration of the Video Motion Consistency Refinement (VMCR) module. The VMCR module minimizes both pixel-wise loss and frequency-wise loss, aligning motion predictions between frames to enhance visual consistency and temporal smoothness in the generated video.}
   \label{fig:freq}
\end{figure}


\subsection{Video Motion Consistency Refinement}
This VMCR module leverages spectral motion analysis to align the predicted latent motion vectors at each diffusion step. By incorporating both global and local motion characteristics from the frequency domain, VMCR ensures smoother motion transitions and temporal coherence across frames in the video generation process.

\vspace{-10pt}

\paragraph{Motion Vector Definition}

$  \hat{z}_0^{(i)}(t) $ representing the $i$-th frame of the predicted denoised output at step  $t$. The residual motion between consecutive frames is captured by the motion difference vector $\delta \hat{z}_t^{i}$, defined as:
\begin{equation}
   \delta \hat{z}_t^{i}:= \hat{z}_0^{(i+1)}(t) - \hat{z}_0^{(i)}(t) ,
\end{equation}

This residual vector  $\delta \hat{z}_t^{i}$ encapsulates the relative motion information between adjacent frames, forming the basis for the alignment and refinement process.

\vspace{-10pt}

\paragraph{Objective Function }
It combines pixel-wise loss $ \ell_{\text{pixel}}$ and frequency-wise loss  $ \ell_{\text{freq}}$ . The total motion alignment loss is formulated as:
\begin{align*}
\ell_{\text{motion}}(\boldsymbol{z}_t) &= \ell_{\text{pixel}}+ \lambda_f \ell_{\text{freq}},
\end{align*}
where $  \lambda_f $  is a hyperparameter controlling the contribution of the frequency-wise loss.
\vspace{-10pt}
\paragraph{Pixel-wise Loss}
The pixel-wise loss \( \ell_{\text{pixel}} \) combines $\ell_2$ loss and cosine similarity loss to capture fine-grained spatial discrepancies between neighbor motion vectors:
\begin{equation}
    \ell_{\text{pixel}} = \sum_{i=0}^{N-2} \left(
     (1 - \cos(\delta \hat{z}_t^{i}, \delta \hat{z}_t^{i+1})) + \lambda_{\text{mse}}
    \| \delta \hat{z}_t^i - \delta \hat{z}_t^{i+1} \|_2^2   \right),
\end{equation}
where $ N $ denotes the number of frames in the video, and $\delta \hat{z}_t^i$  is the $i$-th denoised motion vector derived at step $  t $ , and $  \lambda_{\text{mse}} $ balances the L2 and cosine similarity losses. 
\vspace{-10pt}

\paragraph{Frequency-wise Loss}
The frequency-wise loss \( \ell_{\text{freq}} \) addresses discrepancies in the frequency domain by introducing amplitude and phase losses, defined as:
\begin{equation}
\begin{cases}
\ell_{\text{freq}} &= \ell_{\text{amplitude}} + \lambda_{\text{phase}} \times \ell_{\text{phase}}, \\
\ell_{\text{amplitude}} &= \sum_{i=0}^{N-2} \left| \left| \mathcal{F}\left( \delta \hat{z}_t^{i} \right) \right| - \left| \mathcal{F} \left( \delta \hat{z}_t^{i+1} \right) \right| \right|_1, \\
\ell_{\text{phase}} &= \sum_{i=0}^{N-2} \left| \angle \mathcal{F}\left( \delta \hat{z}_t^{i} \right) - \angle \mathcal{F} \left( \delta \hat{z}_t^{i+1} \right) \right|_1,
\end{cases}
\end{equation}

 After frequency alignment, both the amplitude and phase spectra of different frames converge, ensuring consistent intensity distribution and temporal synchronization of motion across frames.  

\vspace{-10pt}

\paragraph{Optimization of Latent}

At each diffusion step $t $, the predicted latent vector $\boldsymbol{z}_t^i $ is refined through an iterative optimization process aimed at minimizing the total motion loss $ \ell_{\text{motion}} $, defined as:
\begin{equation}
\boldsymbol{z}_t \leftarrow \boldsymbol{z}_t - \omega_{\text{motion}}  \nabla_{\boldsymbol{z}_t} \ell_{\text{motion}}(\boldsymbol{z}_t),
\end{equation}
where $ \omega_{\text{motion}} $ represents  the weight of the VMCR gradient descent. This iterative refinement process ensures alignment of both global and local motion dynamics, contributing to smoother and more temporally coherent video sequences.

\begin{table*}[t]
\centering
\caption{\textbf{Quantitative comparison results}. We compared our GLC Diffusion method against Direct Sampling, FreeLong, GenL, and FreeNoise. Our GLC Diffusion achieves the highest scores in both video quality metrics and temporal consistency metrics (\%), demonstrating its effectiveness in producing high-fidelity and consistent long videos.}
\vspace{-2pt}
\resizebox{\linewidth}{!}{
\begin{tabular}{l|cccccccccc}
\toprule
\multirow{2}{*}{Method} & \multicolumn{5}{c}{\textbf{Video Length $\times$ 3}} & \multicolumn{5}{c}{\textbf{Video Length $\times$ 6}} \\
\cmidrule(lr){2-6} \cmidrule(lr){7-11}
& Sub $\uparrow$ & Back $\uparrow$ & Motion $\uparrow$ & Flicker $\uparrow$ & Imaging $\uparrow$ & Sub $\uparrow$ & Back $\uparrow$ & Motion $\uparrow$ & Flicker $\uparrow$ & Imaging $\uparrow$ \\
\midrule
Direct Sampling & 92.16   & 96.30 &  98.19  & 96.32 & 57.11
& 91.12 & 95.49 & 97.71  & 94.15 & 53.12  \\
FreeLong \cite{lu2024freelong} & 92.72  & 96.72 & 98.10  & 96.60 & 62.73 & 91.96 & 95.58 & 97.75 & 94.19 & 58.90 \\
GenL \cite{wang2023gen} & 91.16 & 94.35 & 98.36 & 96.81 & 66.94 & 89.66 &  93.65 & 98.35 & 96.84 & 66.62 \\
FreeNoise \cite{qiu2024freenoise} & 95.28 & 96.26 & 98.30 & 96.69 & 68.64 & 93.83 &  95.57 & 98.09 & 96.40 & 67.13\\
\midrule
\textit{Ours (GLCD)} & \textbf{95.78} & \textbf{96.92} & \textbf{98.42} & \textbf{96.92} & \textbf{69.86} & \textbf{94.17} & \textbf{95.85} & \textbf{98.35} & \textbf{96.86} & \textbf{68.25} \\
\bottomrule
\end{tabular}
}
\vspace{-10pt}

\label{tab:compare_cog}
\end{table*}


\begin{figure*}[t]
\centering
\includegraphics[width=0.96\linewidth]{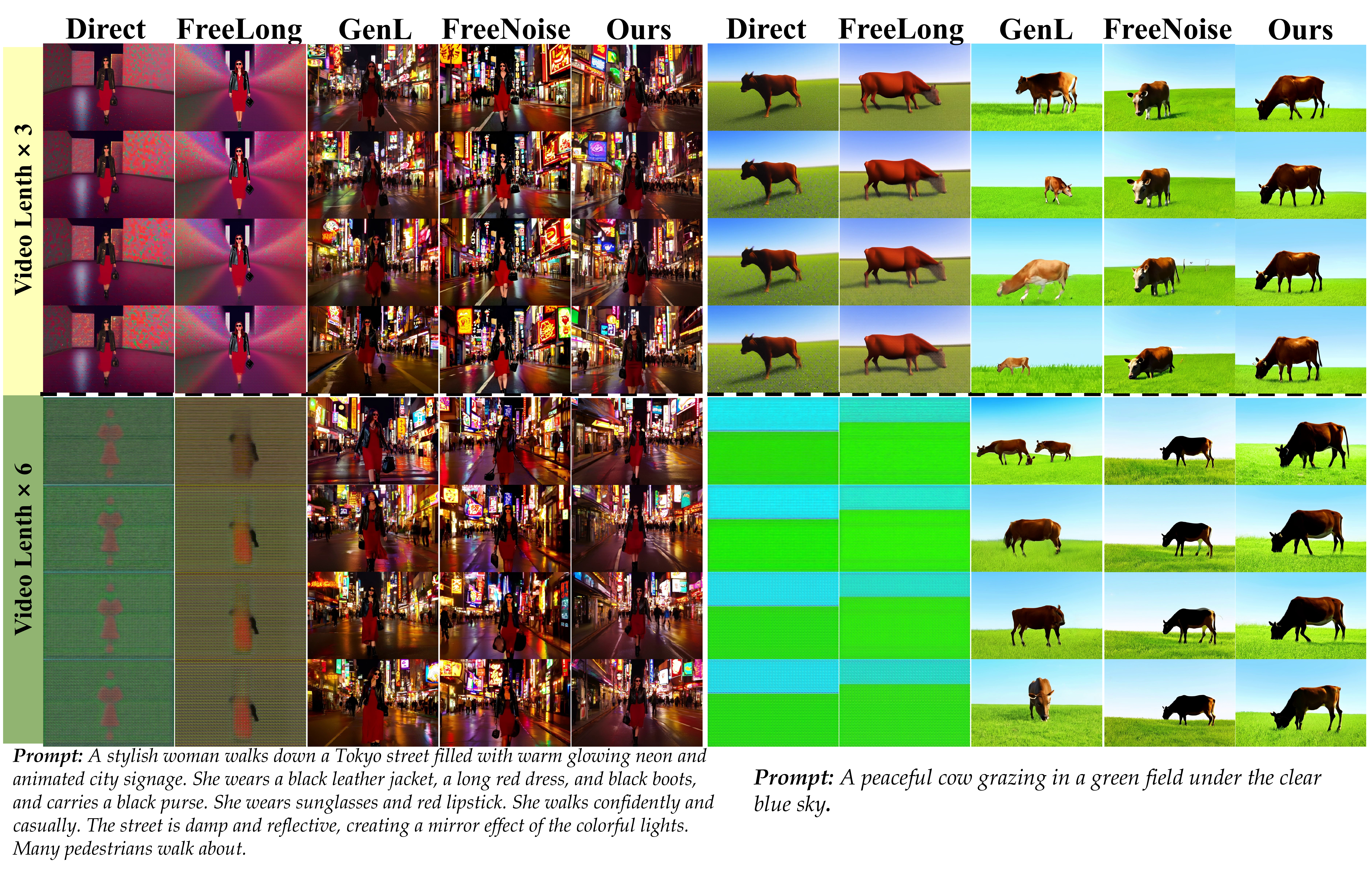}
\vspace{-6pt}
\caption{
Qualitative comparison of long video generation methods with varying lengths (3× and 6×). Visual comparisons are presented for Direct Sampling, FreeLong, GenL, and FreeNoise in order. Direct Sampling and FreeLong produce overly smooth videos with noticeable quality degradation, especially for 6× length, where the visual quality is poor and details are lost. GenL and FreeNoise show improvements in temporal coherence but still suffer from artifacts and significant detail loss. In contrast, our GLC Diffusion consistently generates high-quality videos with smooth motion and consistent content across both 3× and 6× lengths, effectively preserving crucial details and textures.
}

\label{fig:exp_comparison}
\end{figure*}

\vspace{-3pt}

\section{Experiment}
\label{sec:exp}


In this chapter, we report qualitative and quantitative experiments as well as ablation studies. Additionally, we include detailed discussions on hyperparameter settings and empirical experiments in the appendix. The hyperparameters include the annealing coefficient \(\gamma\) in GLCD, the fusion weight \(\lambda\) for ABAM, the frequency loss weight \(\lambda_f\) in the VMCR module, and the gradient descent weight \(\omega_{\text{motion}}\). For more detailed configurations and results, please refer to the appendix.

\subsection{Implement Details}
To evaluate  the effectiveness and generalization capacity of our proposed method, we implement it on the text-to-video generation model CogVideoX \cite{yang2024cogvideox}. Originally, this model is limited to producing videos with a fixed length of 49 frames. We have enhanced the model's capability to generate longer videos. In our experiments, we generated 3$\times$ and 6$\times$ longer videos for quality evaluation across all methods. The evaluations  conducted on a single NVIDIA A800 GPU with a batch size of 1.

\textbf{Evaluation Metrics}:
We utilize VBench to evaluate the quality of long video generation \cite{huang2024vbench}. VBench is designed to comprehensively evaluate T2V models across 16 dimensions, with each dimension tailored to specific prompts and evaluation methods. 
We evaluate video consistency and fidelity using five key metrics selected based on exciting methods \cite{wang2023gen,qiu2024freenoise,lu2024freelong}. For video consistency, we assess subject and background coherence across frames. Video fidelity is measured by analyzing motion smoothness, temporal stability, and overall image quality. These metrics provide a comprehensive assessment of our method's effectiveness in producing high-quality, consistent long video sequences. All metrics are first calculated for each video and then averaged across all videos.

\vspace{-3pt}

\subsection{Quantitative Comparison}
We conducted comparative experiments of our GLC-Diffusion with other tuning-free long video generation methods in diffusion models. (1) Direct Sampling: It involves direct sampling of frames from short VDMs. (2) Gen-L-Video (GenL) \cite{wang2023gen}: It extends to generate smoothly long video by merging overlapping video clips. (3) FreeNoise \cite{qiu2024freenoise}: It aims to enhance temporal coherence across extended noise sequences. (4) FreeLong \cite{lu2024freelong}: It employs the SpectralBlend Temporal Attention mechanism to fuse global low-frequency features with local high-frequency features, enhancing the consistency and fidelity of long video generation.

As shown in Table~\ref{tab:compare_cog}, we selected 200 prompts from VBench \cite{huang2024vbench} to evaluate the effectiveness of our proposed method.
Direct Sampling increases the video duration but leads to a loss of video details and results in noticeable quality degradation. FreeLong improves content consistency by blending global and local features. However, due to the use of identity mapping functions in its global and local paths, it lacks the ability to effectively capture long-range temporal dependencies, leading to temporal inconsistencies in longer videos.  GenL performs well in capturing the semantics of prompts; however, global content variations during the sampling process, caused by overlapping prompts, lead to reduced similarity between consecutive frames. FreeNoise generates relatively stable video results; however, it struggles to produce dynamic scenes and lacks motion variation.
Our GLC-Diffusion results demonstrate superior temporal coherence and content consistency compared to all other methods. We attain the highest scores across all metrics, generating consistent long videos with high fidelity and smooth motion dynamics. By modeling the denoising process as a unified optimization problem and effectively integrating global and local denoising paths with specialized mapping functions, GLC-Diffusion overcomes the limitations of previous methods, including FreeLong, in handling long-range temporal dependencies and maintaining visual quality over extended video sequences.

\begin{table}[t]
\centering
\vspace{-2pt}
    \caption{
    Ablation Study of key components in our  GLC Diffusion. We report metrics related to video quality and temporal consistency.
    }
\resizebox{\linewidth}{!}{%
    \begin{tabular}{l|ccccc}
    \toprule
    Methods & Sub ($\uparrow$) & Back ($\uparrow$) & Motion ($\uparrow$) & Flicker ($\uparrow$) & Imaging ($\uparrow$) \\
    \midrule
    w/o GLCD & 94.53 & 95.78 & 95.06 & 96.22 & 55.30  \\
    \multicolumn{1}{l|}{\hspace{1em}– w/o global} & 94.04 & 95.80 & 98.40 & 96.77 & 67.85 \\ 
    \multicolumn{1}{l|}{\hspace{1em}– w/o local}  & 87.35 & 92.38 & 94.92 & 93.03 & 66.94 \\ 
    w/o Noise Reinit & 95.34 & 96.29 & 97.63 & 95.51 & 69.71 \\ 
    w/o VMCR & 94.41 & 96.06 & 97.78 & 95.72 & 67.83   \\
    \midrule
    \textit{Ours} & \textbf{95.78} & \textbf{96.92} & \textbf{98.42} & \textbf{96.92} & \textbf{69.86}   \\
    \bottomrule
    \end{tabular}
    }

\label{tab:ablation}
\vspace{-6pt}

\end{table}

\begin{figure}[t]
\centering
\includegraphics[width=\linewidth]{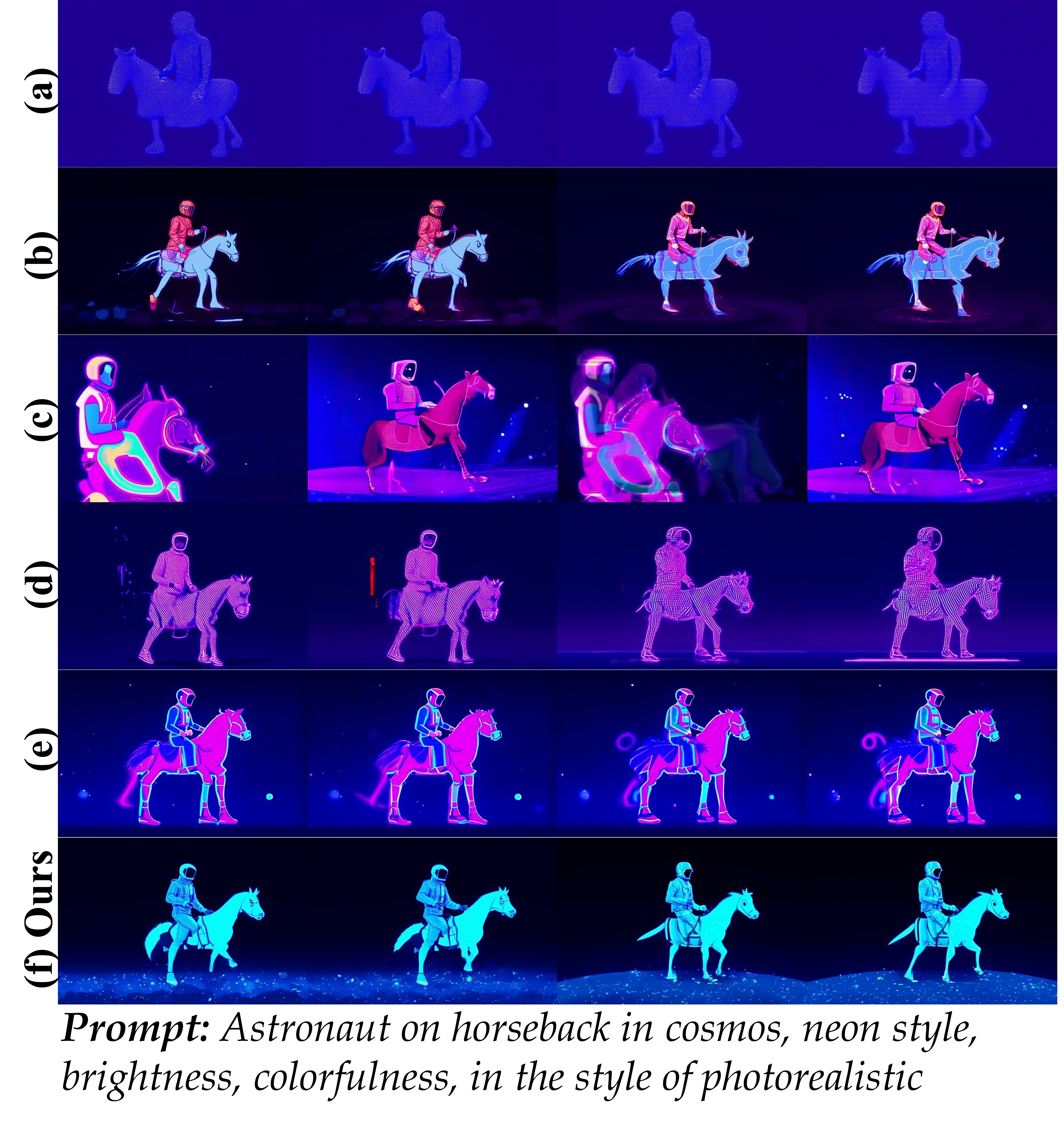}
\vspace{-6pt}
\caption{
Ablation Study on GLC Diffusion Components. We analyze the impact of each component in our method by conducting ablation experiments: (a) w/o GLCD, (b) w/o global path (c) w/o local path, (d) w/o Noise Reinit, (e) w/o VMCR, and (f) Ours. 
}
\label{fig:exp_ablation}
\vspace{-3pt}
\end{figure}


\subsection{Qualitative Comparison}
Our pre-trained model, CogVideoX, generates videos with 49 frames. To test scalability, we extended the video lengths by factors of ×3 and ×6, respectively.
We compared our method with four baselines: Direct, FreeLong, Gen-L-Video, and FreeNoise. The Direct Sampling method, which samples extended frames using models trained on 48-frame sequences, produces poor-quality videos with blurred subjects and unclear backgrounds due to high-frequency distortions. FreeLong relies on single-trajectory denoising and depends heavily on the original model's limited capacity to capture long-term dependencies, resulting in videos that lack temporal consistency. Gen-L-Video attempts longer videos but often results in blurry backgrounds and lacks detail sharpness, failing to maintain scene consistency. FreeNoise enhances global consistency by repeating and shuffling initial noise but fails to generate dynamic scenes with coherent motion. 
In contrast, our method enforces both global and local constraints during the denoising process. By integrating Global Dilated Sampling and Local Random Shifting Sampling within a unified optimization framework, we effectively capture long-range temporal dependencies and enhance spatio-temporal coherence. This enables our model to generate longer videos that maintain high fidelity across frames while accurately reflecting the content described in the prompts.

As illustrated in Figure~\ref{fig:exp_comparison}, our method outperforms all baselines. For the first prompt, our video consistently depicts the stylish woman with sharp details and vibrant, dynamic backgrounds. For the second prompt, our video shows the cow grazing peacefully with smooth motion and consistent scenery. These results confirm the effectiveness of our approach in generating high-quality, temporally coherent long videos that adhere to the given prompts, outperforming existing methods in both visual fidelity and consistency.

\vspace{-3pt}

\subsection{Ablation Studies}
\vspace{-3pt}
To validate the effectiveness of each module in our method, we evaluated three variants: w/o GLCD, w/o Noise Reinit, and  w/o VMCR. Specifically, the w/o GLCD variant further includes w/o global path and w/o local path ablations. 
As shown in Table~\ref{tab:ablation}, the quantitative results demonstrate the significant impact of each module, highlighting their contributions to improving long video generation quality. Specifically, our full method achieves an Imaging score of 69.86, showing improvements over the variants without each component: an increase of 14.56 over w/o GLCD,  2.01 over w/o global path, 2.92 over w/o local path, 0.15 over w/o Noise Reinitialization, and 2.03 over w/o VMCR . These improvements highlight the significant impact of each module on enhancing long video generation quality.

As illustrated in Figure~\ref{fig:exp_ablation}, results indicate that the absence of GLCD leads to inconsistent global content, preventing the video from maintaining a coherent theme and narrative, thereby highlighting GLCD’s role in enforcing overall consistency through the global denoising path and enhancing inter-frame transitions with the local denoising path in overlapping regions. Additionally, removing Noise Reinit module resulted in decreased global content consistency and introduced artifacts during the denoising process, which compromised the visual quality of the video.  Furthermore, without VMCR, videos exhibited reduced motion smoothness and continuity between adjacent frames, underscoring VMCR's importance in maintaining fluid motion and frame-to-frame coherence. This demonstrates that Noise Reinit contributes to global coherence while introducing necessary high-frequency randomness to enhance motion variability. Collectively, these ablation experiments confirm the design rationale behind each module and underscore their combined importance for generating high-quality, temporally consistent videos.

\vspace{-3pt}

\section{Conclusion}
\label{sec:conclusion}
\vspace{-3pt}

In conclusion, we have introduced GLC-Diffusion, a tuning-free method for long video generation that addresses spatiotemporal inconsistencies in existing video diffusion models. By modeling the denoising process as a unified optimization problem through our Global-Local Collaborative Denoising (GLCD) mechanism, we integrate global and local denoising paths to enhance content consistency and temporal coherence. Additionally, the Noise Reinitialization strategy and Video Motion Consistency Refinement (VMCR) module further improve visual consistency and smoothness. Extensive experiments demonstrate that GLC-Diffusion seamlessly integrates with existing models and significantly outperforms previous approaches, marking a substantial advancement in tuning-free long video generation.

\vspace{-6pt}

\paragraph{Limitation}
Our GLC-Diffusion may struggle with scenes involving highly complex or abrupt motion, where accurate alignment across frames becomes challenging, potentially affecting temporal coherence.

{
    \small
    \bibliographystyle{ieeenat_fullname}
    \bibliography{main}
}

\clearpage
\setcounter{page}{1}
\maketitlesupplementary

In this appendix, we provide the following materials:

Sec.~\ref{sec:Algorithm} Algorithm: The denoising process in GLC Diffusion is thoroughly outlined, detailing the steps involved in global-local collaborative denoising, noise reinitialization, and video motion consistency refinement. This algorithm serves as the core foundation of our proposed method.

Sec.~\ref{sec:Hyperparameters} Hyperparameter Settings: A comprehensive summary of the hyperparameters used in each module is provided, including their roles, default values, and chosen ranges for the experiments. These settings illustrate how key parameters such as \(\gamma_0\), \(\lambda\), \(\lambda_f\), and \(\omega_{\text{motion}}\) are tuned to achieve optimal video quality and temporal coherence.

Sec.~\ref{sec:qualitative_results} More Qualitative Results: We present additional qualitative experimental results to validate the effectiveness of our approach. This section includes parameter ablation studies, comparative experiments on videos of different durations, and module ablation analyses.

\begin{algorithm*}[!h]
\caption{Denoising Process in GLC Diffusion}
\label{alg:denoising}
\begin{algorithmic}[1]
\Require 
Text prompt $y$, total timesteps $T$, total video frames $K$
\Ensure 
Generated video frames $\boldsymbol{z}_0$.

\State \textbf{Step 1: Noise Reinitialization}
    \State Randomly sample a local noise unit \(\epsilon \sim \mathcal{N}(0, I)\) and a global noise \(\eta \sim \mathcal{N}(0, I)\)
    
    \State $\boldsymbol{z}_T = \text{NoiseShuffle}(\epsilon)$ 
    \State $\boldsymbol{z}'_T = \text{FreqFusion}(\boldsymbol{z}_T, \eta)$ \Comment{Initialize latent}
    
\For{$t = T, T-1, \ldots, 1$}
    \State \textbf{Step 2: GLCD}

    \State $F_{\mathrm{global}}^i(\boldsymbol{z}_t) = \boldsymbol{z}_t[s_i + d \cdot j]$ \Comment{Global Dilated Sampling}

    \For{each $\boldsymbol{z}_{t,\mathrm{global}}^i$}
        \State $\boldsymbol{z}_{t-1,\mathrm{global}}^i = \phi(\boldsymbol{z}_{t,\mathrm{global}}^i, t, y)$ \Comment{Denoise with $\phi$}
    \EndFor

    \State $\boldsymbol{Z}_{t-1}^{\mathrm{global}} = [\boldsymbol{z}_{t-1,\mathrm{global}}^0, \ldots, \boldsymbol{z}_{t-1,\mathrm{global}}^{N-1}]$ \Comment{Collect denoised global latent}

    \State $F_{\mathrm{local}}^{i,t}(\boldsymbol{z}_t) = \boldsymbol{z}_t[s_i^t + j]$ \Comment{Local Random Shifting Sampling}

    \For{each $\boldsymbol{z}_{t,\mathrm{local}}^i$}
        \State $\boldsymbol{z}_{t-1,\mathrm{local}}^i = \phi(\boldsymbol{z}_{t,\mathrm{local}}^i, t, y)$ \Comment{Denoise with $\phi$}
    \EndFor

    \State $\boldsymbol{Z}_{t-1}^{\mathrm{local}} = [\boldsymbol{z}_{t-1,\mathrm{local}}^0, \ldots, \boldsymbol{z}_{t-1,\mathrm{local}}^{M-1}]$ \Comment{Collect denoised local latent}

    \State $\boldsymbol{z}_{t-1} = \gamma \cdot \mathcal{T}_{\mathrm{global}}(\boldsymbol{Z}_{t-1}^{\mathrm{global}})$ \Comment{Combine global and local paths}

    \State \textbf{Step 3: VMCR}
    \State $\delta \hat{z}_{t-1}^i = \hat{z}_0^{(i+1)}(t-1) - \hat{z}_0^{(i)}(t-1)$ \Comment{Compute motion vector on $\boldsymbol{z}_{t-1}$}

    \State $\ell_{\text{motion}} = \ell_{\text{pixel}} + \lambda_f \cdot \ell_{\text{freq}}$

    \State $\boldsymbol{z}_{t-1} \leftarrow \boldsymbol{z}_{t-1} - \omega_{\text{motion}} \cdot \nabla_{\boldsymbol{z}_{t-1}} \ell_{\text{motion}}$ \Comment{Update latent}

\EndFor

\State \textbf{Output:} Final denoised video frames $\boldsymbol{z}_0$.

\end{algorithmic}
\end{algorithm*}

\section{Algorithm}
\label{sec:Algorithm}

We further illustrate the synthesis process of long videos in Algorithm~\ref{alg:denoising}. 
First, we initialize the latent variable $\boldsymbol{z}'_T$ using the Noise Reinitialization strategy, which combines local noise shuffling and frequency fusion to enhance motion diversity. For the denoising process, we propose Global-Local Collaborative Denoising (GLCD). Specifically, we compute the global representations $\boldsymbol{Z}_t^{\mathrm{global}}$ by applying global dilated sampling to the latent variable $\boldsymbol{z}_t$, capturing long-term temporal dependencies. At the same time, we compute the local representations $\boldsymbol{Z}_t^{\mathrm{local}}$ using local random shifting sampling, focusing on short-term temporal coherence.
After obtaining the global and local representations, we denoise each video clip using the diffusion model $\phi$, resulting in the denoised global and local representations $\boldsymbol{Z}_{t-1}^{\mathrm{global}}$ and $\boldsymbol{Z}_{t-1}^{\mathrm{local}}$. We then fuse the global and local paths, balancing their contributions with the annealing coefficient $\gamma$, and update the latent variable $\boldsymbol{z}_{t-1}$. Additionally, we apply Video Motion Consistency Refinement (VMCR) to $\boldsymbol{z}_{t-1}$. We compute the motion difference vectors $\delta \hat{z}_{t-1}^i$. Then, we minimize the motion alignment loss $\ell_{\text{motion}}$, which consists of pixel loss $\ell_{\text{pixel}}$ and frequency loss $\ell_{\text{freq}}$. We update the latent variable $\boldsymbol{z}_{t-1}$ through gradient descent. This process iterates over each timestep $t$ from $T$ to $0$, ultimately generating the denoised video frames $\boldsymbol{z}_0$

\vspace{-6pt}
\section{Hyperparameter}
\label{sec:Hyperparameters}

In this section, we provide an extensive analysis of the hyperparameter settings used in our proposed method, along with detailed experimental results to validate their impact. Below, we outline the key hyperparameters and their roles, the chosen ranges for the experiments, and the corresponding results. The Global-Local Collaborative Denoising (GLCD) module is configured with an initial annealing coefficient \(\gamma_0 = 0.005\) and a growth rate \(\beta = 0.0005\), allowing for a gradual transition from local to global contributions during the denoising process. The global sampling interval ensures effective capture of long-range dependencies, while the local clip length focuses on short-term temporal coherence. The Anchor-Based Attention Mechanism (ABAM) utilizes a scaling factor \(\lambda = 0.1\) to balance the influence between the anchor clip and the current frame, ensuring temporal consistency without compromising frame fidelity. In the Video Motion Consistency Refinement (VMCR) module, the frequency loss weight \(\lambda_f = 0.2\) balances pixel-wise loss and Frequency-wise loss. The mse loss weight \(\lambda_{mse} = 0.001\) balances the cosine-similarity loss and mse loss in the spatial domain. The phase loss weight $\lambda_{\text{phase}}=1$ balances amplitude and phase losses in Fourier domain. The gradient descent weight \(\omega_{\text{motion}} = 2e-5\) controls the step size during optimization to enhance motion smoothness. These hyperparameters are fine-tuned to provide optimal video quality and temporal consistency, as demonstrated by the experimental results.

\paragraph{Annealing Coefficient $\gamma$ in GLCD}

The annealing coefficient \(\gamma\) dynamically balances the global and local paths in GLCD over the denoising trajectory. The coefficient varies with the timestep \(t\) as: $\gamma = \gamma_0 \cdot e^{\beta \cdot t}$,  where \(\gamma_0\) is the initial annealing coefficient, and \(\beta\) is the growth rate. We default set \(\beta = 0.0005\). The exponential form allows \(\gamma\) to gradually shift the influence from the local path to the global path as the timestep increases, improving global content consistency while retaining temporal coherence.
A lower \(\gamma_0\) gives more weight to local path, improving temporal coherence but potentially sacrificing global content consistency. Conversely, a higher \(\gamma_0\) favors global path, enhancing global-wide coherence while reducing local smoothness. We default set \(\gamma_0 = 0.005\) to balance global and local paths.

We conducted an ablation study to validate the impact of \(\gamma_0\) on video quality and temporal consistency. As illustrated in Table \ref{tab:gamma_ablation}, it summarizes the results for different initial annealing coefficients \(\gamma_0\). The results highlight the trade-offs between global content consistency and temporal coherence under varying \(\gamma_0\) settings. 
As shown in Table \ref{tab:gamma_ablation}, \(\gamma_0 = 0.005\) achieves the best balance between global and local contributions, leading to optimal video quality and temporal coherence.

\begin{table}[t]
\centering
    \caption{
Ablation Study for Initial Annealing Coefficient \(\gamma_0\) in GLCD
    }
\resizebox{\linewidth}{!}{%
    \begin{tabular}{l|ccccc}
    \toprule
    \(\gamma_0\) & Sub ($\uparrow$) & Back ($\uparrow$) & Motion ($\uparrow$) & Flicker ($\uparrow$) & Imaging ($\uparrow$) \\
    \midrule
    
    0.5 & 95.52 & 96.03 & 98.08 & 97.04 & 59.06  \\
    0.05 & 94.97 & 96.18 & 98.37 & 96.90 & 67.07 \\
    0.0005 & 94.83 & 96.07 & 98.37 & 96.77 & 67.48   \\
    \midrule
    \textbf{0.005} & \textbf{95.78} & \textbf{96.92} & \textbf{98.42} & \textbf{96.92} & \textbf{69.86}   \\
    \bottomrule
    \end{tabular}
    }

\label{tab:gamma_ablation}

\end{table}

\paragraph{Fusion Weight $\lambda$ in ABAM}

The ABAM introduces a scaling factor \(\lambda\) to balance the influence of the anchor clip and the current frame during the denoising process. A lower \(\lambda\) places more emphasis on the current frame, potentially improving frame fidelity but possibly reducing temporal consistency. Conversely, a higher \(\lambda\) gives more weight to the anchor clip, enhancing temporal coherence but possibly sacrificing some frame details. We default set \(\lambda = 0.1\) to achieve a balance between frame quality and temporal smoothness.

We conducted an ablation study to evaluate the impact of \(\lambda\) on video quality and temporal consistency. As illustrated in Table~\ref{tab:abam_ablation} summarizes the results for different values of \(\lambda\). The results indicate that \(\lambda = 0.1\) provides the best trade-off, leading to optimal video quality and temporal coherence.

\begin{table}[t]
\centering
    \caption{
Ablation Study for Fusion Weight  \(\lambda\) in ABAM
    }
\resizebox{\linewidth}{!}{%
    \begin{tabular}{l|ccccc}
    \toprule
    \(\lambda\) & Sub ($\uparrow$) & Back ($\uparrow$) & Motion ($\uparrow$) & Flicker ($\uparrow$) & Imaging ($\uparrow$) \\
    \midrule
    1 & 94.50 & 96.08 & 98.32 & 96.77 & 67.85  \\
    0.01 & 93.99 & 95.72 & 98.30 & 96.69 & 68.00 \\
    \midrule
    \textbf{0.1} & \textbf{95.78} & \textbf{96.92} & \textbf{98.42} & \textbf{96.92} & \textbf{69.86}   \\
    \bottomrule
    \end{tabular}
    }

\label{tab:abam_ablation}

\end{table}

\paragraph{Frequency Loss Weight $\lambda_f$ in VMCR}

VMCR utilizes a loss function that operates in both the spatial and frequency domains. The spatial domain loss addresses pixel-level differences between consecutive frames, promoting visual consistency by reducing discrepancies in the image content. In parallel, the frequency domain loss captures motion vector information by aligning the amplitude and phase components of the frames in the frequency spectrum. This frequency alignment helps to diminish spatial artifacts and enhances overall video quality by ensuring consistent motion patterns across frames.

VMCR incorporates a frequency loss weighted by \(\lambda_f\) to maintain temporal coherence by aligning the amplitude and phase of adjacent frames in the frequency domain. A higher \(\lambda_f\) increases the emphasis on frequency alignment, enhancing temporal consistency but potentially affecting spatial details. A lower \(\lambda_f\) may preserve spatial details but reduce temporal coherence. We default set \(\lambda_f = 0.2\) to balance spatial and temporal qualities.

We conducted an ablation study to assess the impact of \(\lambda_f\) on video quality and temporal consistency. As illustrated in Table \ref{tab:lambda_f}, varying \(\lambda_f\) demonstrates the trade-offs between spatial detail preservation and temporal coherence. The results highlight that \(\lambda_f = 0.2\) achieves a satisfactory balance.

\begin{table}[t]
\centering
    \caption{
Ablation Study for Frequency Loss Weight \(\lambda_f\) in VMCR
    }
\resizebox{\linewidth}{!}{%
    \begin{tabular}{l|ccccc}
    \toprule
    \(\lambda_f\) & Sub ($\uparrow$) & Back ($\uparrow$) & Motion ($\uparrow$) & Flicker ($\uparrow$) & Imaging ($\uparrow$) \\
    \midrule
    2 & 94.69 & 95.97 & 98.34 & 96.72 & 68.15  \\
    0.02 & 94.48 & 95.97 & 98.31 & 96.75 & 67.67 \\
    \midrule
    \textbf{0.2} & \textbf{95.78} & \textbf{96.92} & \textbf{98.42} & \textbf{96.92} & \textbf{69.86}   \\
    \bottomrule
    \end{tabular}
    }

\label{tab:lambda_f}

\end{table}

\begin{table}[t]
\centering
    \caption{
Ablation Study for Frequency Loss Weight \(\omega_{motion}\) in VMCR
    }
\resizebox{\linewidth}{!}{%
    \begin{tabular}{l|ccccc}
    \toprule
    \(\omega_{motion}\) & Sub ($\uparrow$) & Back ($\uparrow$) & Motion ($\uparrow$) & Flicker ($\uparrow$) & Imaging ($\uparrow$) \\
    \midrule
    2e-3 & 94.73 & 95.96 & 98.34 & 96.73 & 67.52 \\

    2e-4 & 94.62 & 96.00 & 98.38 & 96.79 & 68.00  \\
    2e-6 & 94.54 & 96.05 & 98.39 & 96.85 & 67.71 \\
    \midrule
    \textbf{2e-5} & \textbf{95.78} & \textbf{96.92} & \textbf{98.42} & \textbf{96.92} & \textbf{69.86}   \\
    \bottomrule
    \end{tabular}
    }
\vspace{-6pt}
\label{tab:w_ablation}

\end{table}

\paragraph{Gradient Descent Weight  $\omega_{motion}$ in VMCR}


In the VMCR module, the gradient descent weight \(\omega_{\text{motion}}\) determines the update step size for motion refinement during the optimization process. A larger \(\omega_{\text{motion}}\) leads to more aggressive updates, which may improve temporal coherence but risk overshooting and introducing artifacts. A smaller \(\omega_{\text{motion}}\) results in more conservative updates, preserving frame quality but possibly insufficiently refining motion consistency. We default set \(\omega_{\text{motion}} =  2e-5\) to balance update aggressiveness and stability.

We performed an ablation study to examine the effect of \(\omega_{\text{motion}}\) on video quality and temporal consistency. As illustrated in Table \ref{tab:w_ablation} presents the results for different \(\omega_{\text{motion}}\) values. The findings indicate that \(\omega_{\text{motion}} = 2e-5\) offers the optimal balance between motion refinement effectiveness and video quality preservation.

\section{More Qualitative Results}
\label{sec:qualitative_results}
In this chapter, we report more qualitative experiment results.

\paragraph{Hyperparameter Ablation Qualitative Results}

\begin{figure*}[t]
\centering
\includegraphics[width=\linewidth]{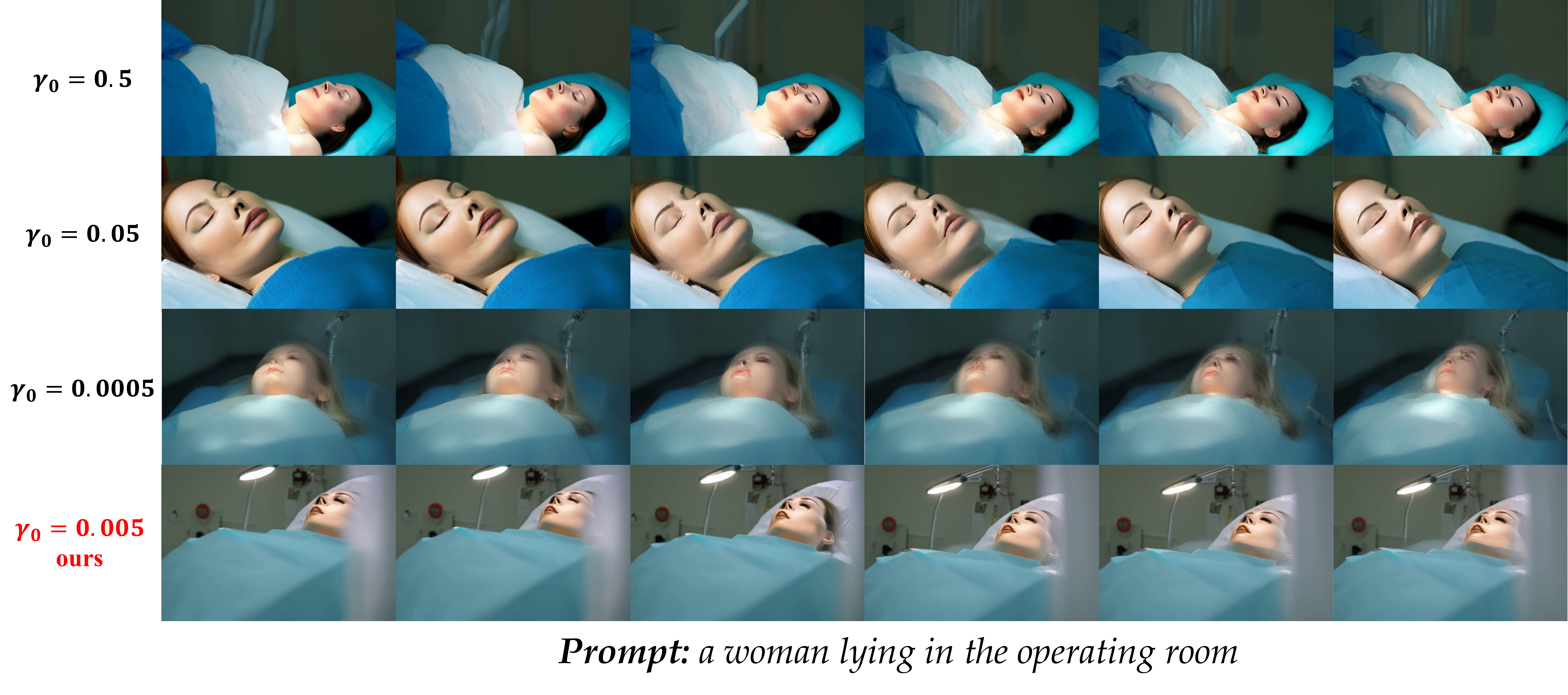}
\caption{
Qualitative Results of Annealing Coefficient $\gamma$ in GLCD.
}
\label{fig:exp_ab_gamma}
\end{figure*}

\begin{figure*}[t]
\centering
\includegraphics[width=\linewidth]{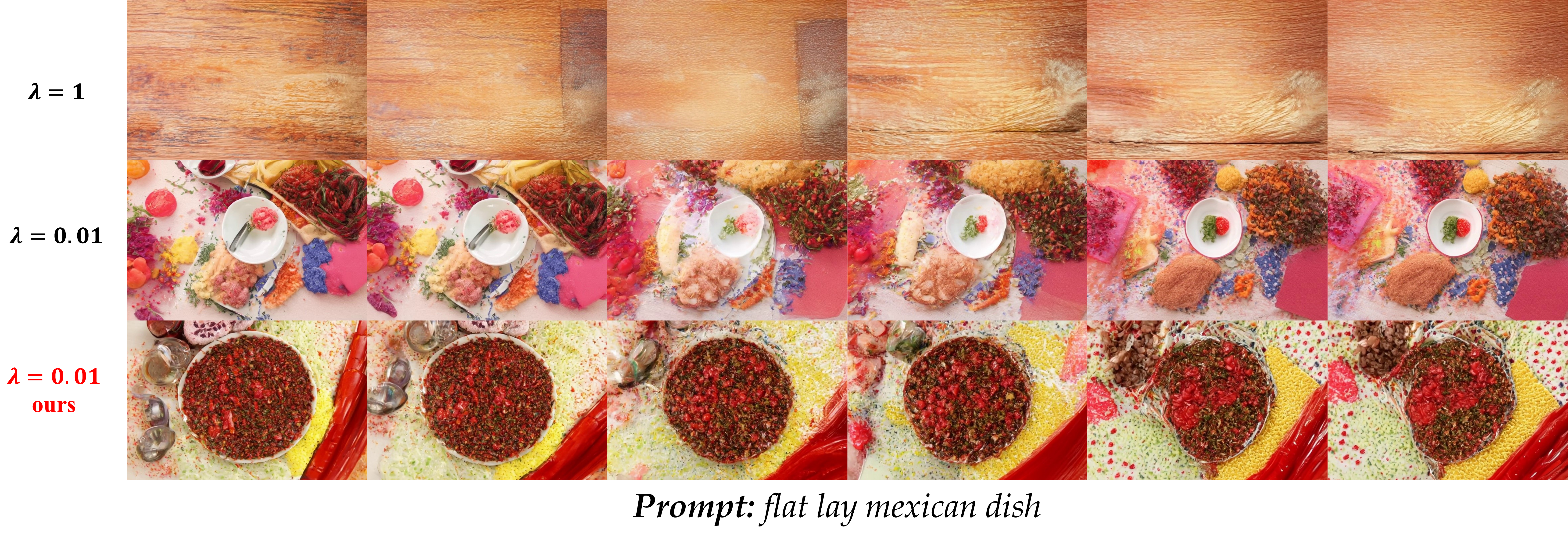}
\caption{
Qualitative Results of Fusion Weight $\lambda$ in ABAM.
}
\label{fig:ab_abam}
\end{figure*}

\begin{figure*}[t]
\centering
\includegraphics[width=\linewidth]{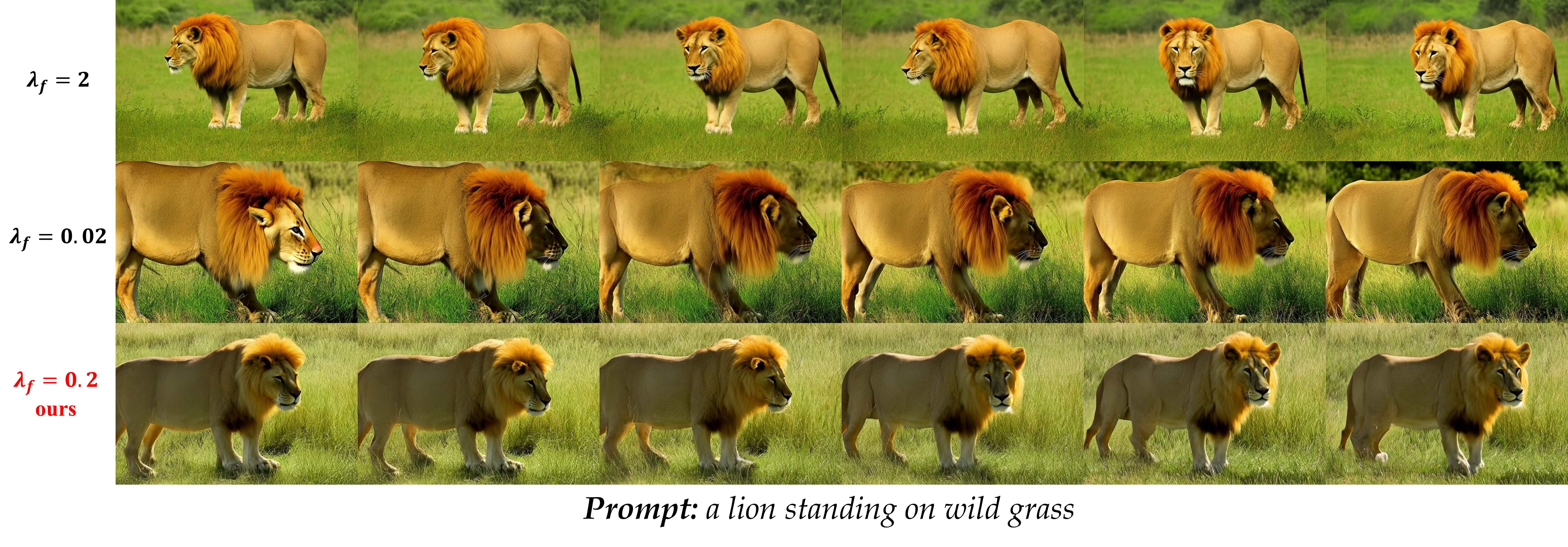}
\caption{
Qualitative Results of Frequency Loss Weight $\lambda_f$ in VMCR.
}
\label{fig:ab_vmcr_loss}
\end{figure*}

\begin{figure*}[t]
\centering
\includegraphics[width=\linewidth]{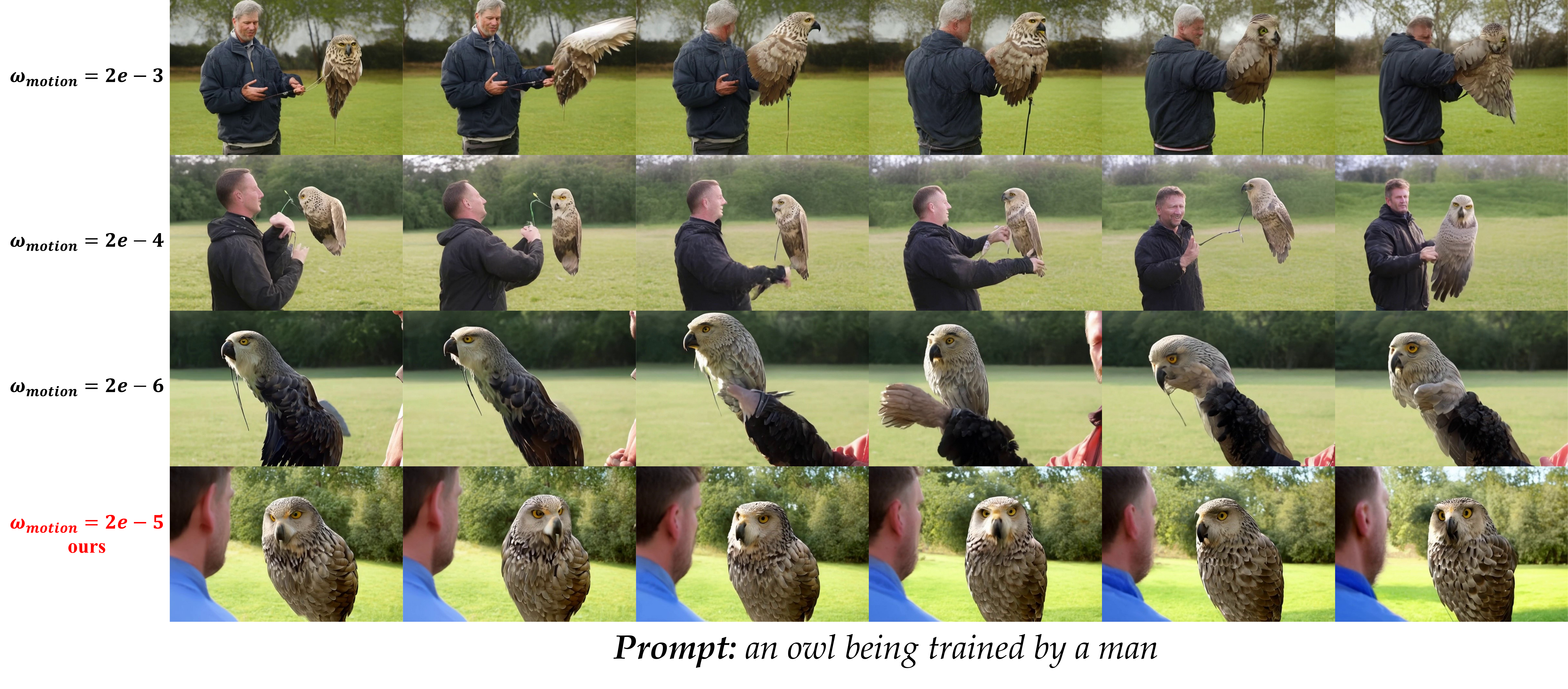}
\caption{
Qualitative Results of Gradient Descent Weight  $\omega_{motion}$ in VMCR.
}
\label{fig:ab_vmcr_w}
\end{figure*}

The qualitative results for different initial annealing coefficients \(\gamma_0\) are illustrated in Figure~\ref{fig:exp_ab_gamma}, highlighting their impact on the outcome. The influence of varying the ABAM scaling factor \(\lambda\) is analyzed, as shown in Figure~\ref{fig:ab_abam}. Comparisons provided in Figure~\ref{fig:ab_vmcr_loss} demonstrate the effect of the frequency loss weight \(\lambda_f\) on motion consistency. The results emphasize its importance in maintaining temporal stability. The gradient descent weight \(\omega_{\text{motion}}\) is shown to influence the update step size during motion refinement in the VMCR module, as depicted in Figure~\ref{fig:ab_vmcr_w}.

\begin{figure*}[t]
\centering
\includegraphics[width=0.96\linewidth]{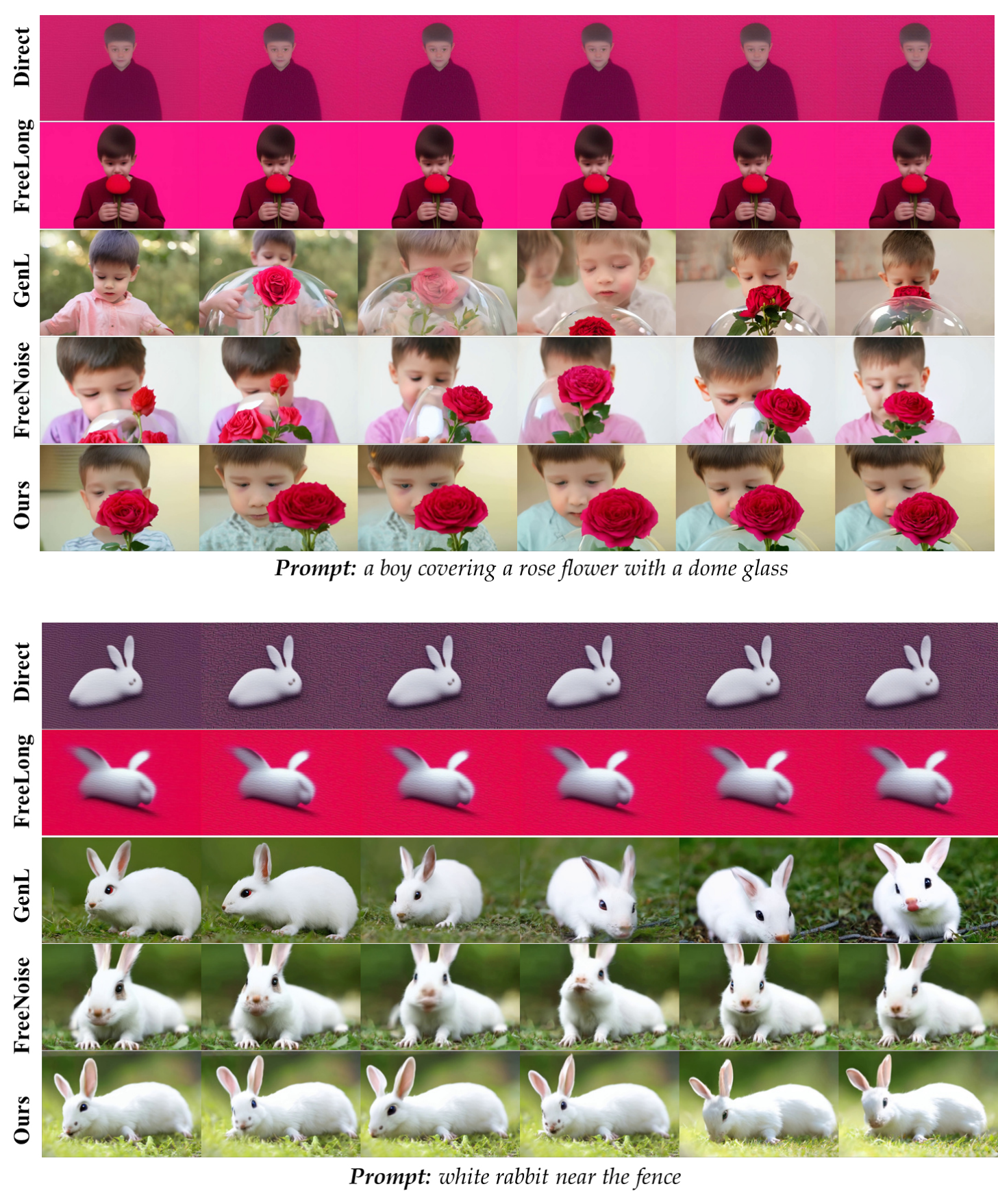}
\caption{
Qualitative comparison of long video generation methods with 3 $\times$ video lengths.
}
\label{fig:exp_comparison_video3}
\end{figure*}

\begin{figure*}[t]
\centering
\includegraphics[width=0.96\linewidth]{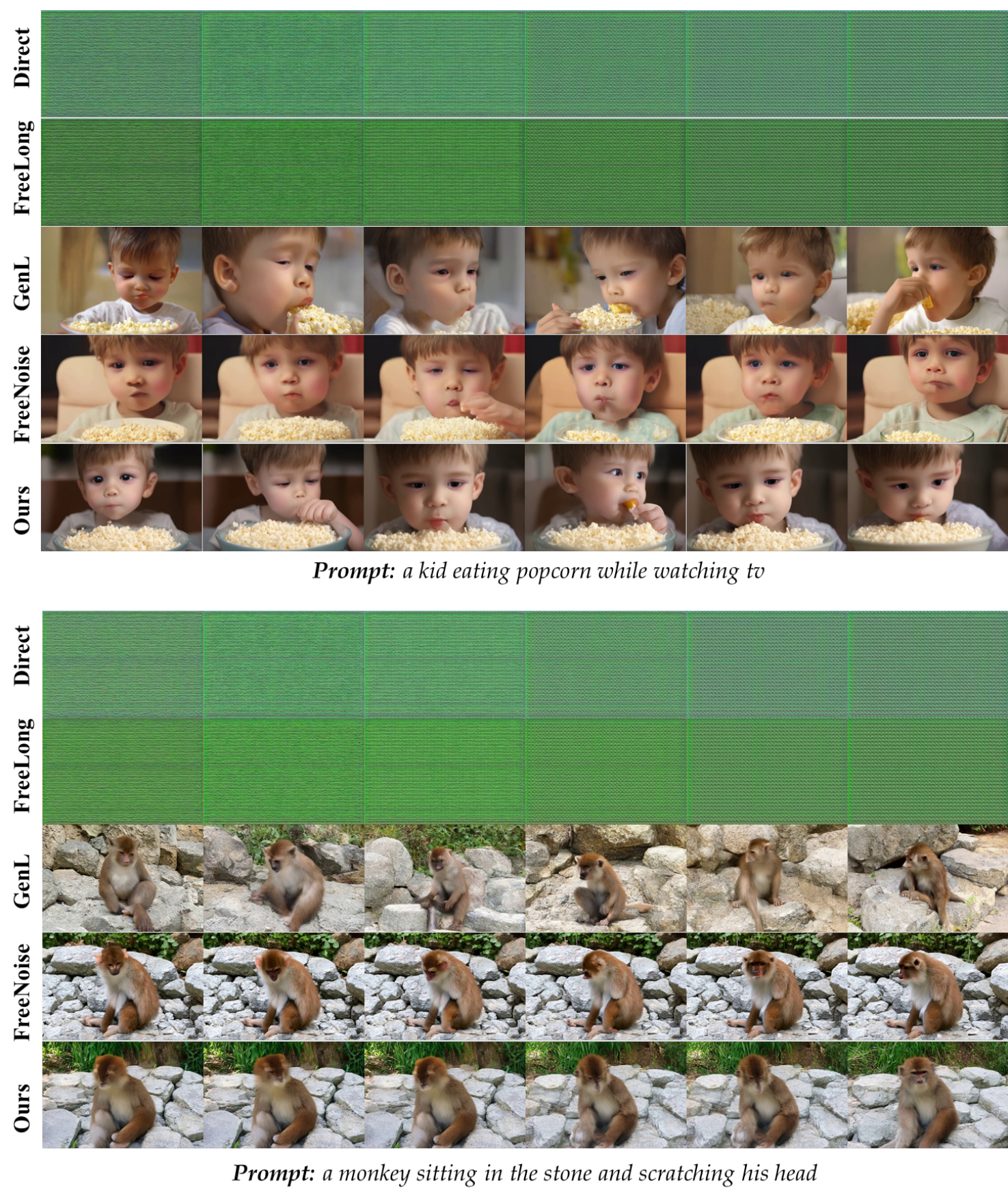}
\caption{
Qualitative comparison of long video generation methods with 6 $\times$ video lengths.
}

\label{fig:exp_comparison_video6}
\end{figure*}

\paragraph{Qualitative Comparsion Results}
To evaluate the scalability and robustness of our method, experiments are conducted on videos of varying lengths. Results comparing videos that are 3 $\times$ and 6 $\times$ longer than the standard duration are presented in Figures~\ref{fig:exp_comparison_video3} and \ref{fig:exp_comparison_video6}. These comparisons highlight the ability of the proposed approach to consistently maintain high visual quality and temporal coherence across different video lengths, demonstrating its effectiveness for long video synthesis.

\paragraph{Module Ablation Qualitative Results}

\begin{figure*}[t]
\centering
\includegraphics[width=0.9\linewidth]{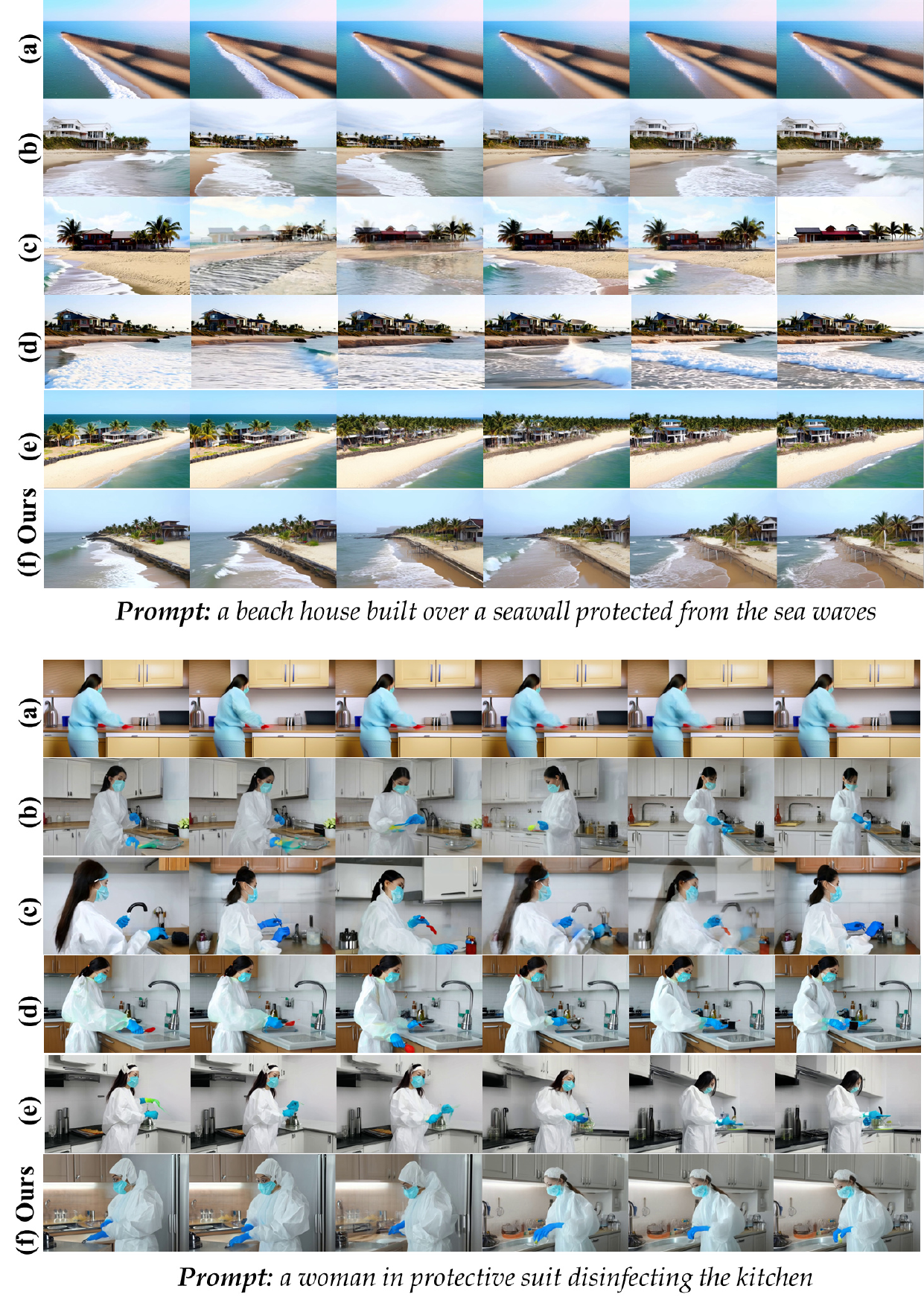}
\caption{
Ablation Study on GLC Diffusion Components: (a) w/o GLCD, (b) w/o global path (c) w/o local path, (d) w/o Noise Reinit, (e) w/o VMCR, and (f) Ours.}
\label{fig:exp_ab_module_1}
\end{figure*}

\begin{figure*}[t]
\centering
\includegraphics[width=0.9\linewidth]{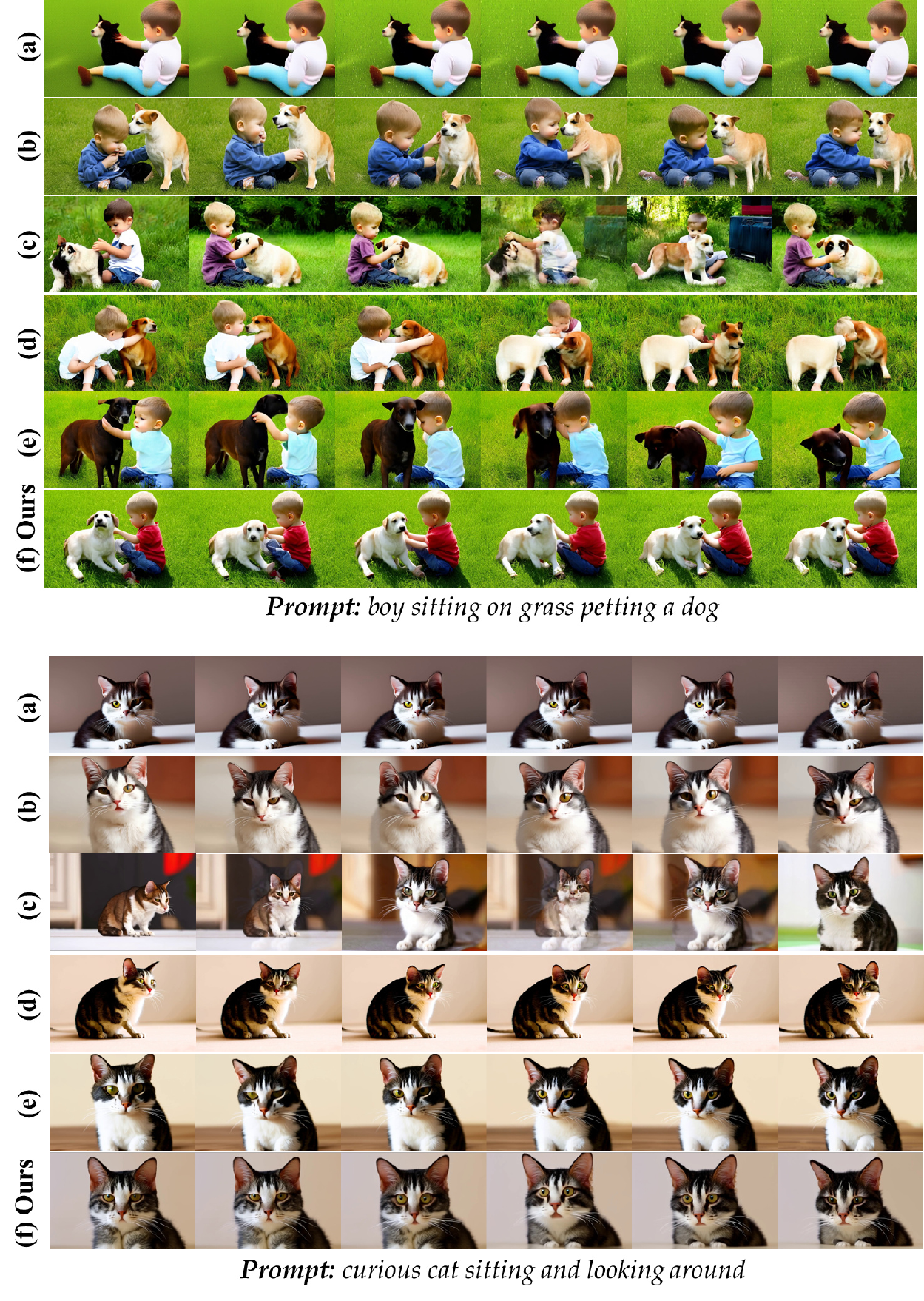}
\caption{
Ablation Study on GLC Diffusion Components: (a) w/o GLCD, (b) w/o global path (c) w/o local path, (d) w/o Noise Reinit, (e) w/o VMCR, and (f) Ours. }
\label{fig:exp_ab_module_2}
\end{figure*}

Additional qualitative comparisons are provided in Figures~\ref{fig:exp_ab_module_1} and \ref{fig:exp_ab_module_2}. These results underscore the significance of each module in contributing to the overall performance of the framework.

\end{document}